\documentclass[runningheads, envcountsame, a4paper]{llncs}
\usepackage[pdftex]{graphicx}

\usepackage{epstopdf}
\usepackage{cite}
\usepackage{amsmath}
\usepackage{multirow}
\usepackage{verbatim}
\usepackage{diagbox}
\usepackage{microtype}
\usepackage{url}

\usepackage{todonotes}

\begin{document}
\title{Automatic Remaining Useful Life Estimation Framework with Embedded Convolutional LSTM as the Backbone} 
%
%
\author{Yexu Zhou \and Yuting Gao \and Yiran Huang \and Michael Hefenbrock \and Till Riedel \and Michael Beigl}
\authorrunning{Y. Zhou et al.}
%
\institute{Telecooperation Office, Karlsruhe Institute of Technology, Karlsruhe, Germany \\ 
\email{$\{$zhou,gao,yhuang,hefenbrock,riedel,michael$\}$@teco.edu}} 
\titlerunning{Automatic Remaining Useful Life Estimation Framework}%
\maketitle              
\begin{abstract}
An essential task in predictive maintenance is the prediction of the Remaining Useful Life (RUL) through the analysis of multivariate time series. Using the sliding window method, Convolutional Neural Network (CNN) and conventional Recurrent Neural Network (RNN) approaches have produced impressive results on this matter, due to their ability to learn optimized features. However, sequence information is only partially modeled by CNN approaches. Due to the flatten mechanism in conventional RNNs, like Long Short Term Memories (LSTM), the temporal information within the window is not fully preserved. To exploit the multi-level temporal information, many approaches are proposed which combine CNN and RNN models. In this work, we propose a new LSTM variant called embedded convolutional LSTM (ECLSTM). In ECLSTM a group of different 1D convolutions is embedded into the LSTM structure. Through this, the temporal information is preserved between and within windows. Since the hyper-parameters of models require careful tuning, we also propose an automated prediction framework based on the Bayesian optimization with hyperband optimizer, which allows for efficient optimization of the network architecture. Finally, we show the superiority of our proposed ECLSTM approach over the state-of-the-art approaches on several widely used benchmark data sets for RUL Estimation.

\end{abstract}
\section{Introduction}
As system complexity and efficiency requirements continue to increase, the strategy of machine maintenance has changed. Where in the past, breakdown corrective maintenance or scheduled preventive maintenance was the standard, now, more intelligent approaches, like predictive maintenance (PM), are strived for. Unlike previous maintenance strategies, PM uses the machine's historical time series sensor data to evaluate the condition. The goal is to proactively maintain the machines before failures occur and therefore minimize down-times. One critical part of PM is the estimation of the remaining useful life (RUL). By quantifying the remaining time until a component loses functionality, downtimes and costs of premature maintenance can be avoided by replacing only components that will fail soon.

Over the past decade, deep learning (DL) has achieved remarkable results in this task. By reviewing the related works, we found that there are still some challenges.

\begin{itemize}
  \item Temporal information preservation and utilization in state-of-the-art algorithms: Recently, researchers focused mainly on constructing various types of neural networks for RUL. However, we found that in many works, the flatten mechanism (layer) is applied. This obfuscates temporal information and potentially leads to under-utilization (This will be further explained in the Section~\ref{sec:related}). While recent works, such as \cite{fusion1,fusion2}, build complex structures to fully preserve and utilize this information, they are often not usable for RUL estimation because of the high model complexity.
  \item Flexibility: As RUL estimation is used for various components with different degradation patterns, such as lithium batteries \cite{lithium1}, Rolling Bearing \cite{bearing} and complex power generation systems \cite{cmapss}, experts have designed specialized deep learning model structures and often employed task-specific feature engineering. However, the design of the model architecture and the setting of hyper-parameters may be challenging for non-experts. Thus a universal automatic prediction framework is of huge benefit for practical applications.
\end{itemize}

To preserve and exploit the multi-level temporal information, we design a novel LSTM variant called embedded convolutional LSTM (ECLSTM). Instead of prepending a convolutional layer, it allows inputs of each time step to be 2-dimensional or 3-dimensional. The convolution can make full use of local temporal information and the complexity of the model does not depend on the size of the window but only the width of the kernel. In order to address the difficulty of setting the hyper-parameters of the model, we propose an automatic deep learning framework for RUL estimation. In the framework, we apply stacked ECLSTM as the backbone to extract features. By using Bayesian Optimization and Hyperband (BOHB) \cite{BOHB}, this framework can automatically adapt all hyper-parameters involving in the whole data analysis pipeline without expert knowledge. We validate the performance of our approach on a number of real-world public RUL data sets. The results show, that the proposed ECLSTM has a superior prediction ability over competing state-of-the-art approaches.

\section{Related Work} \label{sec:related}
The methods of RUL estimation can be roughly categorized into model-based methods and data-driven methods. Deriving a model-based method is difficult, as it requires an accurate understanding of the underlying physical phenomena. Data-driven methods, on the other hand, model the degradation characteristic based on historical sensor data. Among the data-driven methods, DL-based approaches attract the most attention due to their ability to learn task-specific features from time series.

\begin{figure}
\hspace{-0.1cm}
\includegraphics[width=\textwidth]{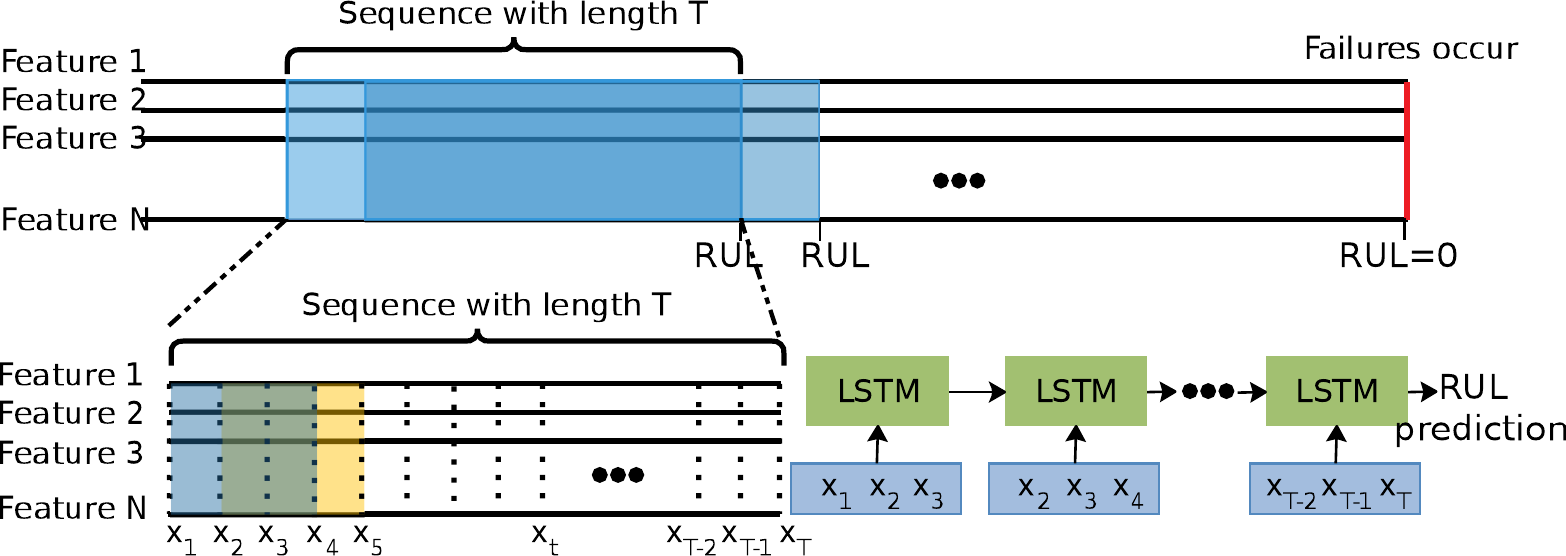}
\caption{The sliding window method to turn a multivariate time series data into a supervised learning problem (top). Features in each labeled window are sequentially fed into the (unrolled) LSTM (bottom right). } 
\label{fig:sliding}
\end{figure}

In order to train DL models, the time series problem is reframed as a supervised regression problem using the sliding window method (see Fig.~\ref{fig:sliding}). Time series data is segmented into overlapping, fixed-length sequences where each sequence is assigned a label (the RUL). In RUL estimation tasks, the aim is to predict the corresponding RUL at a given point in time.

\begin{figure}
\hspace{-0.1cm}
\includegraphics[width=\textwidth]{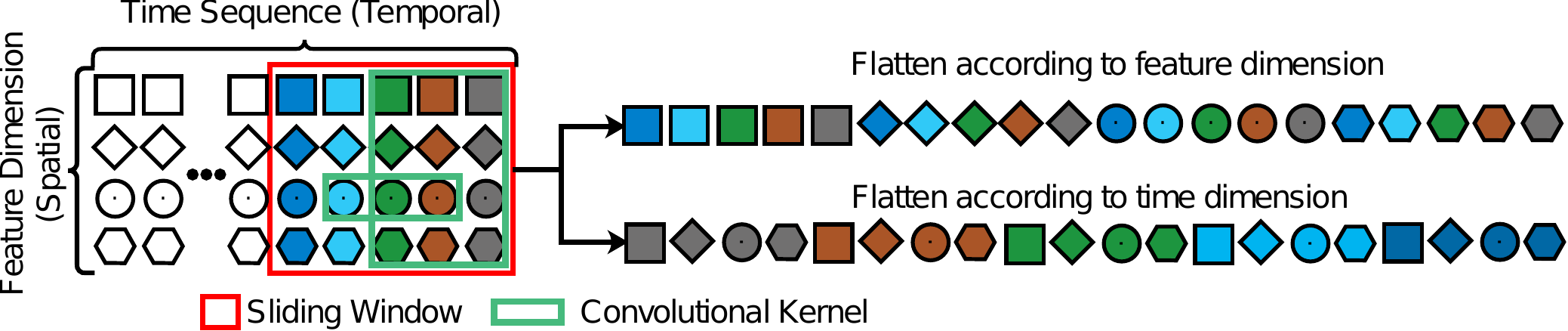}
\caption{Multiple features (shapes) at multiple sampling times (colors) within one window must be flattened into a 1D vector for LSTM input destroying the natural sequence. Alternatively, a convolutional kernel can be applied to aggregate local information.} \label{fig:reshape}
\end{figure}

In order to predict the RUL, early works applied simple RNNs \cite{RNN}. However, to address the problem of learning long-term time dependencies in RNN, LSTM models \cite{LSTM1,LSTM2} consume the values at each point in time sequentially. In order to provide more context and enhance the feature extraction, the sliding window method (see Fig.~\ref{fig:sliding}) can be used \cite{LSTMwindow}. Here, all values in the window are fully connected to an LSTM layer. Due to the fully connected layer in the LSTM (FCLSTM), the input of each time step must be one-dimensional, so that temporal and feature structure need to be projected (see Fig.~\ref{fig:reshape}). Furthermore, the model complexity (number of weights) increases linearly with the window size, which is unfavorable.

Alternatively Convolutional Neural Network (CNN) based methods \cite{DCNN1, DCNN2}, however, they can not perceive temporal patterns exceeding the window size. Therefore, in order to combine the strengths of shared weights in CNNs with an additional long term memory capability, \cite{CNNLSTM1, CNNLSTM2} combined LSTMs and CNNs. In such architectures, the flattening problem persists. Late fusion strategy may be applied to deal with this problem \cite{fusion1,fusion2}, which adds model complexity and may thus easily lead to over-fitting. 

The structure of Convolutional LSTM (ConvLSTM) \cite{convlstm} is widely used in video processing in which the spatial information in images and sequential information in videos are fully preserved. Inspired from that, we design the ECLSTM with a more general structure targeted at multivariate time series.

Automated machine learning is now a very popular research direction. There are many open-source tools such as auto-sklearn \cite{autosklearn} and auto-weka \cite{autoweka}. These tools help users automatically select the best model and the best hyper-parameters. Based on the existing automated machine learning framework, we build a tool that can automatically perform RUL estimation. The implementation of the framework is available online\footnote{\url{https://github.com/YexuZhou/AutoRUL} }.
\section{Problem Definition}
By reviewing various open-source data sets, we consider the RUL estimation problem in a more general form that takes round-robin sampling strategy into account. In actual production or run-to-failure experiments, the sensor information is not always recorded as storing large amounts of high-precision floating-point data is expensive and requires extremely high bandwidth to transmit. Therefore, the data is typically collected using a round-robin sampling strategy, which records values only once per predefined cycle. For example, in the C-MAPSS data set \cite{cmapss}, only one sample is recorded in each cycle, while in the FEMTO-ST bearing data set \cite{femto}, values are recorded at a frequency of 25.6KHZ \mbox{0.1 seconds} per each cycle (\mbox{10 seconds}). 

Formally, RUL estimation can be described as a sequence to target problem. Given a ${T}$ length sequence time series ${{\bf{X}} = \left({{\bf{x}}_t} \mid {t = 1,\cdots,T} \right)}$ with ${{{\bf{x}}_t} \in {R^{n \times m}}}$, where ${n}$ is the number of sensors and ${m}$ the number of samples per cycle. Now, the aim is to predict the corresponding output ${{y_T}}$, ${{y_T} = f\left({{\bf{x}}_t}\mid  {t = 1, \cdots ,T} \right.)}$, where ${{{\bf{x}}_t}}$ denotes all samples at cycle ${t}$. When the sliding window method is applied, the previous formula should be modified to ${{y_T} = f({\bf{x}}_t^w\left| {t = w,\cdots,T} \right.)}$, where ${w}$ is the size of the window. The vector ${{\bf{x}}_t^w}$ thereby contains all the samples in the time window which is denoted as ${{\bf{x}}_t^w = ({{\bf{x}}_{t - w + 1}}, \cdots ,{{\bf{x}}_t})}$. In our settings, the size of sliding step is always set to 1. For any model, the sequence length determines how past information is used, while the window size describes the complexity of dynamic features over time. As both parameters can greatly affect the model performance, both should be considered when optimizing the model.
\section{Embedded Convolutional LSTM}
 Inspired by the work ConvLSTM \cite{convlstm}, we propose an extension of FCLSTM, in which a group of different 1D convolutions is embedded into the LSTM structure, which we call Embedded Convolutional LSTM (ECLSTM) . We assume that such ECLSTM architecture is more powerful than FCLSTM in handling multivariate time series tasks.
 
 In order to preserve the temporal information within the window, the input should be kept as a 2-dimensional tensor. This can be achieved by replacing the full connection in the FCLSTM with the convolutional operation. The equations of ECLSTM are then given by
\begin{equation} \label{eq:conv}
\begin{array}{l}
{i_t} = \sigma ({W_i}*[{x_t},{h_{t - 1}}] + {b_i})\\
{f_t} = \sigma ({W_f}*[{x_t},{h_{t - 1}}] + {b_f})\\
{o_t} = \sigma ({W_o}*[{x_t},{h_{t - 1}}] + {b_o})\\
{C_t} = {f_t} \circ {C_{t - 1}} + {i_t} \circ \tanh ({W_C}*[{x_t},{h_{t - 1}}] + {b_C})\\
{h_t} = {o_t} \circ \tanh ({C_t}),
\end{array} 
\end{equation}
where ${*}$ indicates the convolution operator and ${\circ}$ the element-wise product. There are three benefits to using the convolution operator in LSTM. Firstly, the convolution parameters are only related to the defined kernel size and the number of filters and not to the size of the window. When the window size is large, the complexity of the model does not increase with it. Secondly, the hidden state $H$ and the memory $C$ also become 2D tensors. This means that they implicitly inherit and preserve the temporal relationship. Thirdly, the input, hidden state, and memory can even maintain a 3-dimensional shape, as this will not affect the operation of the convolution. Keeping the three-dimensional shape allows for more different convolutions. 

The stacking of convolutional layers allows a hierarchical decomposition of the raw data and combinations of lower-level features. In order to get more complex features, the convolutions in \eqref{eq:conv} can be stacked as convolutional cells in a chain structure. If three convolutional layers are stacked in the cell, we call the ECLSTM as 3-depth-ECLSTM. Taking the input gate in ECLSTM as an example, the activation can be calculated as
\begin{equation}
{i_t} = \sigma (W_i^3*\sigma (W_i^2*\sigma (W_i^1*[{x_t},{h_{t - 1}}] + b_i^1) + b_i^2) + b_i^3).
\end{equation}
Other gates have the same structure, but do not share the weights.

\begin{figure}
\hspace{-0.1cm}
\includegraphics[width=\textwidth]{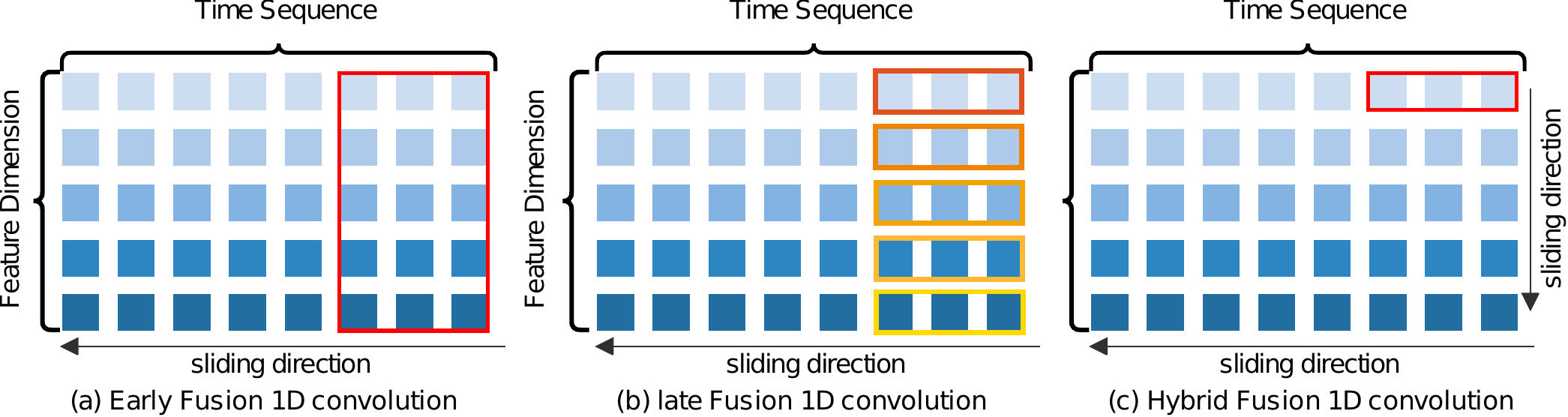}
\caption{In the early fusion convolution, the kernel height is fixed, that is, the same as the number of features. The sliding direction of the convolution kernel is along the time axis. In the late fusion convolution, the kernel height is 1. Each feature has its own convolution kernel. The convolution kernel also has only one sliding direction, namely the time axis. In hybrid fusion convolution, the kernel height is also 1. But it has two sliding directions, one is the time axis and the other is the feature axis. Because of weight sharing, it can save many parameters. It should be noted that when the number of filters is greater than 1, the output of the early fusion convolution is 2-dimensional. The outputs of the two remaining convolutions are 3-dimensional.} \label{fig:fusion}
\end{figure}

Moreover, the results of many multivariate time series analysis works like  \cite{fusion2} indicate that different fusions strategies affect the performance. Inspired by that, the convolution cell can be composed of the following three different 1-dimensional convolutions, which are shown in the Fig.~\ref{fig:fusion}. The first is the early fusion convolution, which is same as conventional 1D convolution. Here, the features are extracted from all sensory information jointly. The second is the late fusion convolution. In late fusion, the features are extracted separately from each sensor. The third is hybrid fusion convolution, where features are separately extracted from each sensor but weights are shared. 

\section{Automatic Prediction Framework}
Because the above architecture is sensitive to hyper-parameters that are difficult to choose even by domain experts, we introduce the automatic prediction framework. Our assumption is that such a framework will outperform typical state of the art time series prediction algorithms when applied to RUL prediction.

The framework's structure is designed based on a summary of the structures from other related works that use neural networks to handle multivariate time series tasks. This framework consists of three parts, namely the pre-processing, feature extraction and RUL prediction parts. The structure and configuration space of each part will be introduced separately. Then the optimization process of the entire framework will be shown at the end of this section.

\subsubsection{Pre-processing}
\begin{figure}
\hspace{0cm}
\includegraphics[width=\textwidth]{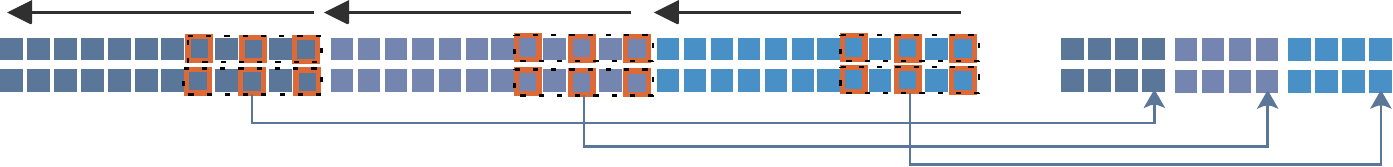}
\caption{Convolutions applied to a window size of 3 over 3 cycles. Each color represents a cycle of 12 samples. All convolutions are performed simultaneously in each cycle and feature with shared weights. A kernel width of 3, the dilation rate of 2, and the stride of 2 reduce the dimensionality to 4 samples per cycle each feature. } \label{fig:downsample}
\end{figure}

When the sampling frequency in one cycle is too high, it is impractical to directly use the raw data as the input to the recurrent neural network. Processing so many values requires a larger kernel width or a relatively deep convolutional network. Both will lead to increased model complexity i.e. parameters and lead to over-fitting.  Traditional methods to reduce the mode complexity are feature extraction or down-sampling. Since down-sampling leads to a loss of information, and manual feature extraction needs to be done on a per task basis, we avoid both for our framework. In order to control the model complexity while still being able to extract sophisticated features, we resort to convolutions with a kernel height of 1. Through this, the dimensionality can be reduced while useful features can be extracted automatically (illustration see Fig.~\ref{fig:downsample}). 
By adjusting the kernel width, the stride, and the dilation rate, the intensity of the dimensional reduction can be adjusted. Note that in this step the conventional 1D convolution is applied and the weights used by the convolution in each cycle and each feature are shared.

The configuration space for this section is shown in Table~\ref{tab:Configuration}. There are dependencies between parameters. If the number of layers is defined as 0, it means that no pre-processing is performed. Only when the number of levels is greater than 0, other parameters are used. These parameters are not defined separately for each layer. As a rule of thumb, we define simple rules here. All layers are initialized with the same number of filters, stride and dilation rate. The kernel width is halved in each subsequent layer. Additionally, whether to perform 1d max-pooling between layers can be selected.

\subsubsection{Feature Extraction}

The main part of a neural network model is the backbone network, which is responsible for extracting features. Our framework uses stacked ECLSTM as the backbone. By default, batch normalization is applied between layers. Table~\ref{tab:Configuration} also shows the configuration space for this part. We define the hyper-parameters on a per-layer basis.  Although this is a relatively simple setup, it guarantees the diversity of the backbone network where subsequent layers can have different structures. Due to the requirement for the optimized time budget, the training time for each configuration is not expected to be too long. Therefore, we limit the maximum number of stacked ECLSTM layers to 4. After the number of stacked layers is determined, the depth of the convolution needs to be defined for each layer. Finally, the type of convolution, the number of filters and the kernel width are set for each convolution.

\subsubsection{RUL Prediction} 
The RUL prediction part is composed of stacked fully connected layers. The input is the feature extracted by the previous backbone. The last layer outputs the final prediction. As shown in Table~\ref{tab:Configuration}, each layer needs to define the number of nodes, the activation function, and the size of the dropout.

\begin{table*}[t]
\caption{Configuration space of the proposed framework.}
\label{tab:Configuration}
\centering
\begin{tabular}{|l|l|l|l|l|} 
\hline
Part & Conditioned on   & Name   & Range   & Type    \\
\hline
\hline
\multirow{9}{*}{Pre-processing} &  -  & no. of layers     & $ \left[0,5\right] $& int                  \\
\cline{2-5}
            &\multirow{6}{*}{no. of layers}     & kernel width         & $\left[ 2,1024 \right]$& int                   \\
\cline{3-5}
 &                                & strides              & $\left[ 1,20 \right]$  & int                    \\
\cline{3-5}
    &                               & dilation rate        & $\left[ 1,10 \right]$  & int                    \\
    
\cline{3-5}                     &                                & activation & \begin{tabular}[c]{@{}l@{}}\{"Sigmoid","ReLU",\\"Leaky","Linear",\\"hard sigmoid"\}\end{tabular} & cat                  \\
\cline{3-5}
                     &                                & no. of filters    & $\left[ 1,20 \right]$  & int                   \\
\cline{3-5}
                     &                                & 1D max pooling    & \{"True","False"\}  & cat                     \\
                     \hline
\multirow{9}{*}{Feature Extraction}          &   -  & no. of layers     & $\left[ 1,3 \right]$     & int                     \\
\cline{2-5}
                     & \multirow{2}{*}{no. of layers} & dropout in layer     & $\left[ 0.0,0.99 \right]$                                                                                   & float                   \\
\cline{3-5}
                     &                                  & depth of cell        & $\left[ 1,4 \right]$                                                                                        & int                     \\
\cline{2-5}
  & \multirow{4}{*}{depth of cell}    & activation & \begin{tabular}[c]{@{}l@{}}\{"Sigmoid","ReLU",\\"Leaky","Linear",\\"hard sigmoid"\}\end{tabular} & cat                    \\
\cline{3-5}
                     &                                 & convolutional type            & \{"early","hybrid","late"\}                                                                    & cat                    \\
\cline{3-5}
                     &                                 & no. of filters    & $\left[ 4,64 \right]$                                                                                       & int                    \\
\cline{3-5}
                     &                                 & kernel width         & $\left[ 2,32 \right]$                                                                                       & int                   \\
\hline
\multirow{6}{*}{RUL Prediction} &   -    & no. of layers & $\left[ 1,4 \right]$ & int   \\
\cline{2-5}
            & \multirow{5}{*}{no. of layers } & no. of units  & $\left[ 8,1024 \right]$  & int   \\
\cline{3-5}
            &  & activation      & \begin{tabular}[c]{@{}l@{}}\{"Sigmoid","ReLU",\\"Leaky","Linear",\\"hard sigmoid"\}\end{tabular} & cat   \\
\cline{3-5}
            &                   & dropout in layer    & $\left[ 0.0,0.99 \right]$  & float   \\
\hline

\end{tabular}
\end{table*}

\subsubsection{Hyper-parameter Optimization}
In addition to the structure parameters introduced earlier, also the sequence length, window size, and training batch size need to be determined. One common solution is the use of random search, but that is very inefficient. For such a highly conditional configuration space, the proposed framework uses the BOHB optimizer \cite{BOHB}, which consists of the Bayesian optimization (BO) and the Hyperband (HB). By repeatedly calling SuccessiveHalving (SH), HB can efficiently identify the best of n randomly-sampled configurations. More time budget will be invested in promising configurations and the configurations with poor performance will be stopped early. BO uses a kernel density estimator (KDE) to model the objective function of configurations. It describes the regions of high performance in the configuration space. In our framework, the objective function is the performance of 3-folds cross validation. 

\section{Experiments} \label{sec:experiments}
In this paper we have stated two hypotheses that we want to support with our experiments: (H1) Our ECLSTM architecture is more powerful than FCLSTM when applied to common multivariate time series prediction tasks and (H2) that when included into an automated prediction framework it can outperform state-of-the-art multivariate time series prediction algorithms on RUL prediction benchmark data sets. Further we want to explore if such an approach might generalize to similar time series prediction problems.

In this section, we evaluate the proposed framework on four benchmark data sets, three from the domain of predictive maintenance and one human activity recognition task. We compare our results to state-of-the-art approaches and perform an ablation study to provide insights into the effectiveness of the different elements of the framework. All experiments are run on a single GPU(RTX 2080 8G RAM). Models are trained with the Adam optimizer \cite{adam}.

\subsection{C-MAPSS Data Set}
 C-MAPSS data set \cite{cmapss} which contains turbofan engine degradation data is a widely used prognostic benchmark data for predicting the RUL. This data set is simulated by the tool Commercial Modular Aero Propulsion System Simulation (C-MAPSS) developed by NASA. Run-to-failure simulations were performed for engines with varying degrees of initial wear but in a healthy state. During each cycle in the simulation, one sample of all 21 sensors such as the physical core speed, the temperature at fan inlet and the pressure at fan inlet etc. will be recorded once. As the simulation progresses, the performance of the turbofan engine degrades until it loses functionality. 
 
\begin{table}
\caption{ Description of four sub-data sets from C-MAPSS. }
\label{tab:CMAPSSData}
\centering
\begin{tabular}{|l|llll|} 
\hline
Data Set               & FD001   & FD002   & FD003   & FD004    \\ 
\hline
Training set           & 100     & 260     & 100     & 249      \\
Test set               & 100     & 259     & 100     & 248      \\
Operational conditions & 1       & 6       & 1       & 6        \\
Fault conditions       & 1       & 1       & 2       & 2        \\
\hline
\end{tabular}
\end{table}

\subsubsection{Data Description} 
C-MAPSS data consists of four sub-data sets with different operational conditions and fault patterns. As shown in Table~\ref{tab:CMAPSSData}, each sub-data set has been split into a training set and a test set. The training sets contain sensor records for all cycles in the run-to-failure simulation. Unlike the training sets, the test sets only contain partial temporal sensor records which stopped at a time prior to the failure. The task is to predict the RUL of each engine in the test sets by using the training sets with the given sensor records. The corresponding RUL to test sets has been provided. With this, the performance of the model can be verified. It should be noted that the four sub-data sets are of varying complexity.

\subsubsection{Evaluation Metrics} 
There are two performance metrics utilized to measure the quality of our proposed model and other benchmark models. One is the commonly used root mean squared error (RMSE) and the other is the scoring function which was used in the PHM08 prognostics challenge competition. RMSE is a symmetric loss that assigns the same penalties for over- and under-prediction. However in practice, over- and under-prediction will lead to different value consequences. The scoring function assigns more penalty when the predicted RUL is larger than the true RUL. Due to being based on such a prediction, the maintenance plan will be delayed. The following equations show these two evaluation metrics.

\begin{equation}\label{eq:rmes}
\mathit{RMSE} = \sqrt {\frac{1}{N}\sum\limits_{i = 1}^N {{{({{\hat r}_i} - {r_i})}^2}} }   
\end{equation}

\begin{equation}\label{eq:score}
\mathit{score} = \left\{ {\begin{array}{*{20}{c}}
{\sum\limits_{i = 1}^N {({e^{ - \frac{{{{\hat r}_i} - {r_i}}}{{13}}}} - 1),\;\;\;\;\;\text{if}} \;\;{{\hat r}_i} < {r_i}}\\
{\sum\limits_{i = 1}^N {({e^{ - \frac{{{{\hat r}_i} - {r_i}}}{{10}}}} - 1),\;\;\;\;\;\text{if}} \;\;{{\hat r}_i} \ge {r_i}}
\end{array}} \right.
\end{equation}
where ${N}$ denotes the total number of engines in the test sets. ${\hat r}$  and  ${{r_i}}$  represent predicted RUL and true RUL respectively.

\subsubsection{Data Preparation} 

The data preparation for the C-MAPSS data set mainly consists of two aspects, one is normalization and the other is the definition of the RUL objective function. In order to do the following ablation study and comparison with other published work, the data preparation work is consistent with the works as \cite{LSTMwindow,CNNLSTM1}. Due to the different distribution of the values from each sensor, these values are normalized through the z-score  normalization method which is described in the following equation ${{{x'}_i} = ({x_i} - {\mu _i})/{\sigma _i}}$. ${\mu _i}$ represents the mean of ${i}$-th sensor and ${\sigma _i}$ the corresponding standard deviation. The RUL objective function assigns each cycle an RUL label. Here we adopt the piece-wise linear RUL target function, where the maximum RUL is defined as 130.

\subsubsection{Ablation Study} 
To validate whether the ECLSTM can improve performance with the sliding window method, in this study, we investigated two fundamentally similar neural network architectures evaluated on the four sub-data sets. One neural network model, called FCLSTM, was taken from \cite{LSTMwindow}, which is often cited as a baseline by other related works. The network has a five-layer architecture. The first layer is a LSTM layer with 32 hidden nodes. The second layer is the same, but contains 64 hidden nodes. The third and fourth layers are forward fully connected layers, each with 8 hidden nodes. The last layer is a 1-dimensional output layer which predicts the RUL. This architecture has been optimized on the four sub-data sets in \cite{LSTMwindow}. The other neural network model is composed of the proposed ECLSTM. The difference from the former network is that the first and second layers are replaced by 2-depth-ECLSTM layer. To determine the hyper-parameters (convolution kernel size and filters number) in ECLSTM layers we apply simple rules. The number of filters in these two layers is simply defined as 10. The size (width) of the convolution kernel varies with the window size, that is, the rounded value of the window size divided by 4.  The window sizes can be 1, 5, 10, 15 and 20. According to the work \cite{LSTMwindow}, the sequence length is defined as 30. These two architectures were trained 10 times for each window size. Their performance on the test set was recorded.

\begin{figure}
\hspace{-0.1cm}
\includegraphics[width=\textwidth]{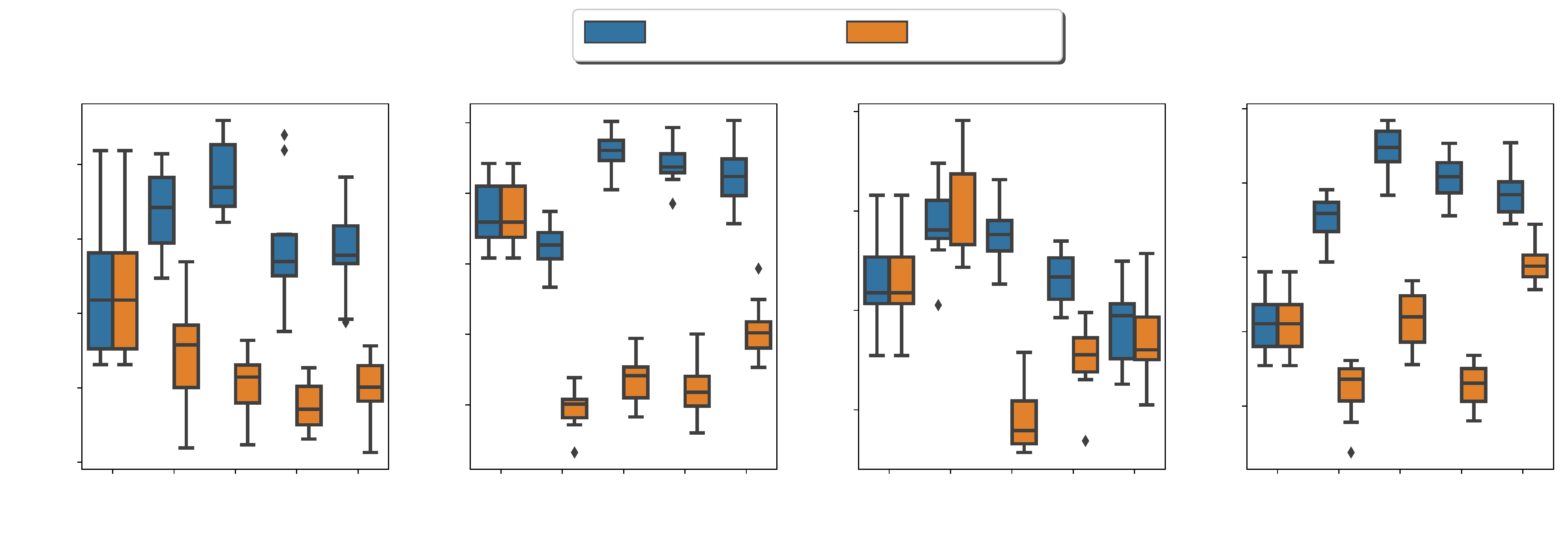}
\caption{Ablation study comparing ECLSTM against FCLSTM models on the four sub-data sets of C-MAPSS using different window sizes on RMSE metric (lower is better).} 
\label{fig:ablation}
\end{figure}

In most cases, the performance of FCLSTM deteriorates with the increase of the window size. This can be seen in Fig.~\ref{fig:ablation}. Only on the FD002 and FD003 sub-data, when the window sizes are equal to 5 and 20 respectively, the performance of FCLSTM improves, but it is not significant. The reason is that with flatten, the temporal relationship within the window is ignored. In contrast when the window size becomes larger, the performance of ECLSTM, in most cases, is improved. We also observe that the performance of ECLSTM keeps increasing from window size equals 1 to 15 but drops at 20. This is likely due to less temporal information between windows. Because the sequence length is fixed, if the window size is too large, the time-steps (number of windows) becomes smaller. Smaller time-steps means less temporal information between windows, which leads to worse model performance.

Here we conducted a significance test and the results show that the performance of ECLSTM compared to FCLSTM with window size 1 is significant. Also, on sub-data FD002, ECLSTM has already achieved the best result compared to other state-of-the-art methods. It is worth noting that the input sensor is not being selected and the structure of the network can be further optimized. Therefore, the ECLSTM's performance still has room for improvement. Because the difference between the two structures lies only in the first and second layers, and ECLSTM achieves better performance on all four sub-data sets. So through this experiment, it proves that when the window size is not equal to 1, ECLSTM can more efficiently extract temporal information.

\subsubsection{Results of Proposed Framework}
To get better results, the proposed framework is used to optimize the architecture of the model. The search space consists of the hyper-parameters listed in Table~\ref{tab:Configuration}. the optimization was run on GPU RTX 2080 for 24 hours. The final optimized model is trained 10 times with different seeds. The average results on the test set are listed in Table~\ref{tab:resultCMAPSSData}. The approaches listed in the table are all recently published. Among them \cite{Adabn} is currently the best performer on C-MAPSS data set. It can be seen that the optimized model achieved the best results on all four sub-data sets, whose RMSE values of the 4 sub-data sets are the lowest. Especially on the FD002 and FD004 sub-data set, the estimation accuracy has improved significantly.

\begin{table}
\caption{ RMSE and Score comparison on C-MAPSS data.}
\label{tab:resultCMAPSSData}
\centering
\begin{tabular}{l|cccccccc}
\hline
Datasets   & \multicolumn{2}{c}{FD001} & \multicolumn{2}{c}{FD002} & \multicolumn{2}{c}{FD003} & \multicolumn{2}{c}{FD004}  \\ 
\hline
\diagbox{Methods}{Metric}   & $\mathit{RMSE}$  & $\mathit{score}$ & $\mathit{RMSE}$ & $\mathit{score}$ & $\mathit{RMSE}$ & $\mathit{score}$ & $\mathit{RMSE}$ & $\mathit{score}$    \\ 
\hline
LSTM \cite{LSTMwindow}  & 16.14 & 3.38x10$^{2}$          & 24.49 & 4.45x10$^{3}$            & 16.18 & 8.52x10$^{2}$            & 28.17 & 5.55x10$^{3}$             \\
DCNN \cite{DCNN1}      & 12.61 & 2.74x10$^{2}$          & 22.36 & 1.04x10$^{5}$            & 12.64 & 2.84x10$^{2}$            & 23.31 & 1.25x10$^{5}$             \\
DAG \cite{CNNLSTM1}       & 11.96 & 2.29x10$^{2}$          & 20.34 & 2.73x10$^{3}$            & 12.46 & 5.35x10$^{2}$           & 22.43 & 3.37x10$^{3}$             \\
AdaBN  \cite{Adabn}    & 11.94 & 2.20x10$^{2}$         & 19.29 & 2.25x10$^{3}$            & 12.31 & 2.60x10$^{2}$            & 22.14 & 3.63x10$^{3}$             \\
Proposed   & \textbf{11.03}  &   \textbf{2.16x10$^{2}$} & \textbf{15.95} & \textbf{1.44x10$^{3}$}         & \textbf{11.23}  & \textbf{1.93x10$^{2}$}         & \textbf{16.21} &     \textbf{1.4x10$^{3}$}               \\ 
\hline
\end{tabular}
\end{table}

\subsection{PHM 2008 Data Set}
This data set is similar to C-MAPSS data set. It was used for the prognostics challenge competition at the International Conference on Prognostics and Health Management (PHM) in the year 2008. It contains only one failure mode and therefore shows lower complexity. Associated true RUL values to test trajectories are not revealed. After the model is trained on the training trajectories, the results on the test trajectories need to be uploaded to the website, which will return a final score value as equation \eqref{eq:score}. Data preparation is the same as it on C-MAPSS data set. We use piece-wise linear RUL as the target function, where the maximal RUL is 130. The values of each sensor are normalized through the z-score normalization. We let our framework optimize for 12 hours and select the model with the best validation performance as the final model. The score results are shown in Table~\ref{tab:PHM08}. 

\begin{table}
\caption{$\mathit{score}$ of prediction results compared to PHM '08 rognostics challenge }
\label{tab:PHM08}
\centering
\begin{tabular}{|l|>{\centering\arraybackslash}p{0.1\textwidth}|>{\centering\arraybackslash}p{0.1\textwidth}|}
\hline
     Methods  & Year  & $\mathit{score}$   \\ 
\hline
     Competition rank 1 & 2008 & 436.841   \\
     Competition rank 2 & 2008 & 512.426    \\
     Competition rank 3 & 2008 & 737.769   \\
     Competition rank 4 & 2008 & 809.757   \\
     \textbf{proposed framework} & 2020 & 823.341  \\
     Competition rank 5  & 2008 & 908.588    \\
     Competition rank 15 & 2008 & 1557.61   \\
     Deep LSTM \cite{LSTMwindow}  & 2017 & 1862     \\
     Deep CNN  \cite{DCNN2} & 2016 & 2056    \\ 
\hline
\end{tabular}
\end{table}

\subsection{FEMTO-ST Bearing Data Set}

\begin{table}
\caption{Data sets of IEEE 2012 PHM Prognostic Challenge.}
\label{tab:PHM2012}
\centering
\begin{tabular}{|l|l|l|lllll} 
\cline{1-3}
Operation Conditions & Training set  & Test set &  &  &  &  &   \\
\cline{1-3}
1800 rpm and 4000 N  & Bearing1\_1,~ Bearing1\_2~ & \begin{tabular}[c]{@{}l@{}}Bearing1\_3,~ Bearing1\_4, Bearing1\_5,\\Bearing1\_6, Bearing1\_7\end{tabular} &  &  &  &  &   \\ 
\cline{1-3}
1650 rpm and 4200 N  & Bearing2\_1,~ Bearing2\_2   & \begin{tabular}[c]{@{}l@{}}Bearing2\_3, Bearing2\_4, Bearing2\_5,\\Bearing2\_6, Bearing2\_7\end{tabular}  &  &  &  &  &   \\ 
\cline{1-3}
1500 rpm and 5000 N  & Bearing3\_1,~ Bearing3\_2   & Bearing3\_3 &  &  &  &  &   \\
\cline{1-3}
\end{tabular}
\end{table}
The FEMTO-ST data set \cite{femto} was used in the PHM Challenge in the year 2012 for the RUL estimation of bearings. As shown in Table~\ref{tab:PHM2012}, run-to-failure experiments were performed on 17 bearings. There were three different conditions for the experiment. In each condition, the data of 2 bearings were used as the training set. The model is trained on 6 bearings and then predicts the RUL of the remaining 11 bearings. Two accelerators are mounted on the outer ring of the bearing and vertical and horizontal vibrations are recorded. The sampling frequency is 25.6KHZ. It Records 0.1 seconds every 10 seconds. That is, 2560 samples are recorded per cycle. To evaluate the performance of methods, the absolute percent error of predicted results are used, which was also applied in the challenge. It is defined as ${E{r_i} = abs(\frac{{ActRU{L_i} - RU{L_i}}}{{ActRU{L_i}}} \times 100\%) }$, where ${{ActRU{L_i}}}$ is the actual RUL and ${{RU{L_i}}}$ the predicted RUL of the ${i}$-th testing bearing.

After the values of the two sensors are normalized by z-score normalization, we let the optimizer BOHB of the framework run for 24 hours. The model with the best validation performance is trained on all training data. its performance on the test set is shown in Table~\ref{tab:femot}. The work in \cite{CCG}  has the best performance among existing studies, in which the original data is first converted into time-frequency images by continuous wavelet transform. Lei Y. et al. \cite{lei} proposed to apply the mutual information from multiple time series to construct the health indicator. Guo L. et al. \cite{guo} applied a set of selected features based on expert knowledge as input to the RNN model. For our framework, no expert knowledge is used for pre-processing and we reach comparable results (see Table~\ref{tab:femot}).

\begin{table}
\caption{FEMTO-ST RUL prediction mean absolute percentage error for each model.}
\label{tab:femot}
\centering
\begin{tabular}{|l|l|l|l|l|l|l|}
\hline
Testing~             & Actual & Predicted          & RNN-HI \cite{guo} & WMQE\cite{lei} & CCG-HI \cite{CCG}             & Proposed              \\ 
\hline
Bearing1\_3          & 5730                 & 5883                 & 43.28\%              & 0.35\%               & 25.09\%              & 2.67\%                \\ 
\hline
Bearing1\_4          & 339                  & 495                  & 67.55\%              & 5.6\%                & 16.22\%              & 46.02\%               \\ 
\hline
Bearing1\_5          & 1610                 & 1357                 & 22.98\%              & 100.00\%             & 15.34\%              & 15.71\%               \\ 
\hline
Bearing1\_6          & 1460                 & 1715                 & 21.23\%              & 28.08\%              & 26.30\%              & 17.47\%               \\ 
\hline
Bearing1\_7          & 7570                 & 7227                 & 17.83\%              & 19.55\%              & 6.68\%               & 4.53\%                \\ 
\hline
Bearing2\_3          & 7530                 & 5810                 & 37.84\%              & 20.19\%              & 31.23\%              & 22.84\%               \\ 
\hline
Bearing2\_4          & 1390                 & 1804                 & 19.42\%              & 8.63\%               & 25.39\%              & 29.78\%               \\ 
\hline
Bearing2\_5          & 3090                 & 3855                 & 54.37\%              & 23.3\%               & 41.65\%              & 24.76\%               \\ 
\hline
Bearing2\_6          & 1290                 & 1510                 & 13.95\%              & 58.91\%              & 11.24\%              & 17.05\%               \\ 
\hline
Bearing2\_7          & 580                  & 670                  & 55.17\%              & 5.17\%               & 12.41\%              & 15.52\%               \\ 
\hline
Bearing3\_1          & 795                  & 950                  & 3.66\%               & 40.24\%              & 3.05\%               & 19.50\%               \\ 
\hline
Mean of Error        &                      &                      & 34.28\%              & 28.18\%              & 18.51\%              & 19.62\%               \\
\hline
\end{tabular}
\end{table}

\subsection{UCI Human Activity Recognition (HAR)}
In order to further verify if the proposed approach may also be applicable to general multivariate time series prediction task aside from RUL estimation, we conduct an experiment on the UCI-HAR data set \cite{har}. HAR has the same problem definition as RUL estimation. Given a fixed length sequence multivariate time series, the aim is to predict the corresponding activity.  This data set consists of sensor signals (accelerometer and gyroscope) gathered from a smartphone by 30 volunteer subjects. The volunteers performed six activities (walking, walking upstairs, walking downstairs, sitting, standing, laying). A total of 10298 sequences were collected. The data has been pre-processed and each sequence has 128 samples. Moreover, UCI-HAR data set provides a train-test partition. 21 volunteers were selected for generating the training data and 30${\%}$ the test data. To evaluate the performance, balanced accuracy is used as the metric. After the values are normalized through z-score normalization, we let the optimization run for 12 hours. The model with the best cross validation performance is trained 10 times with different seeds. The balanced accuracy on the test data set is listed in Table~\ref{tab:har}.
\begin{table}
\caption{Comparison of different models (publication year) using mean balanced accuracy on UCI-HAR data set}
\label{tab:har}
\centering
\begin{tabular}{|c|c|c|} 
\hline
  CNN-LSTM \cite{cnnlstm}  & HCF+CNN \cite{hcf}  & \textbf{Proposed}   \\ 
  (2019) & (2020) & (2020)\\
\hline
$93.40\%$  &  $93.80\%$ & \textbf{$94.69\% \pm 0.42\%$}   \\ 
\hline                       
\end{tabular}
\end{table}

\section{Conclusion and Future Work}
In this work, an automatic RUL estimation framework is proposed. The backbone of the framework is the ECSLTM, which effectively combines the strengths of CNNs and LSTMs. In order to show the effectiveness of the framework, we evaluated its performance on different, real-world benchmark data sets. From our experiments, the following conclusions can be drawn:
\begin{enumerate}
  \item The window size can affect the performance of the model. Compared to LSTMs, the proposed ECLSTM has better performance with an increase in window size.
  \item Our framework has achieved state-of-the-art results on the three benchmark RUL estimation data sets. Especially on the C-MAPSS data set, a significant improvement compared to other recently published methods was achieved. Furthermore, the entire process does not require expert knowledge for architecture and parameter tuning and is therefore user-friendly for non-experts.
  \item To check if the proposed framework works for general multivariate time series tasks, We additionally validate the framework's capabilities on an activity recognition data set. The result shows that our proposed framework achieves state-of-the-art performance. This framework can be applied to other similar multivariate time series tasks.
\end{enumerate}

For future work, the diversity of the framework should be increased through e.g. adding attention mechanisms and including properties from bidirectional ECLSTM. At the same time, we will try to design a more reasonable configuration space. The configuration space defines how the architectures can be represented, which also determines the difficulty of the hyper-parameter optimization problem. A well-designed configuration space can simplify the optimization and may discover better models faster\cite{nas}.

%
%
%
\bibliographystyle{splncs04}
\bibliography{llncs}
\end{document}